\documentclass[runningheads]{llncs}

\usepackage{graphicx}

\usepackage{cite}
\usepackage{amsmath,amssymb,amsfonts}
\usepackage{algorithmic}
\usepackage{graphicx}
\usepackage{textcomp}
\usepackage{xcolor}
\usepackage{url}
\usepackage{multirow}

\begin{document}

\title{Gates Are Not What You Need in RNNs\thanks{We would like to thank the Faculty of Computing, University of Latvia for covering the costs related to the conference, the IMCS UL Scientific Cloud for the computing power, and Leo Trukšāns for the technical support. This research is funded by the Latvian Council of Science, project No.~lzp-2018/1-0327 and lzp-2021/1-0479.}}

%
%
\author{Ronalds Zakovskis\inst{1}\thanks{Corresponding author.} \and
Andis Draguns\inst{2} \and
Eliza Gaile\inst{1} \and
Emils Ozolins\inst{2} \and
Karlis Freivalds\inst{3}}
%
\authorrunning{R. Zakovskis et al.}
%
\institute{Faculty of Computing, University of Latvia, Riga, Latvia\\
\email{ronalds.zakovskis@gmail.com}\\
\email{eliiza.gaile@gmail.com} \and
Institute of Mathematics and Computer Science at University of Latvia, Riga, Latvia\\
\email{andis.draguns@lumii.lv}\\
\email{ozolinsemils@gmail.com} \and
Institute of Electronics and Computer Science, Riga, Latvia
\email{karlis.freivalds@edi.lv}}

\maketitle              

\begin{abstract}
Recurrent neural networks have flourished in many areas. Consequently, we can see new RNN cells being developed continuously, usually by creating or using gates in a new, original way. But what if we told you that gates in RNNs are redundant? In this paper, we propose a new recurrent cell called \textit{Residual Recurrent Unit} (RRU) which beats traditional cells and does not employ a single gate. It is based on the residual shortcut connection, linear transformations, ReLU, and normalization. To evaluate our cell's effectiveness, we compare its performance against the widely-used GRU and LSTM cells and the recently proposed Mogrifier LSTM on several tasks including, polyphonic music modeling, language modeling, and sentiment analysis. Our experiments show that RRU outperforms the traditional gated units on most of these tasks. Also, it has better robustness to parameter selection, allowing immediate application in new tasks without much tuning. We have implemented the RRU in TensorFlow, and the code is made available at \url{https://github.com/LUMII-Syslab/RRU}.

\end{abstract}

\section{Introduction}
\label{section:introduction}

Recurrent neural networks (RNN) have achieved widespread use in sequence processing tasks such as language modeling, speech and music recognition.
RNNs are composed of a single computation cell, called Recurrent Unit which is unrolled along the sequence dimension.

In order to achieve stable training and the ability to store and make use of long-term dependencies, the recurrent units are designed in a special way. The most well-known units are LSTM and GRU, which use a gating mechanism to store and extract information from the recurrent state. Two kinds of gates are used in these units -- reset gates and update gates, where each of them is based on modulating the state information with the input information and is implemented by multiplying two values, one of which is often range-limited by the sigmoid function. Gates are considered crucial for the propagation of long-term information \cite{greff2016lstm} and virtually all proposals of Recurrent Units, including numerous recent developments, contain such multiplicative input-state interactions.
Recent developments of improved RNN variants contain even more multiplicative interactions \cite{krause2016multiplicative} that culminate in Mogrifier LSTM, where the input information is modulated through a series of reset gates \cite{melis2019mogrifier}. They find that about five reset gates achieve the best performance yielding Mogrifier LSTM to be the top-performing RNN cell on several datasets so far. 

\begin{figure}[t]
\begin{center}
\includegraphics[width=0.95\linewidth]{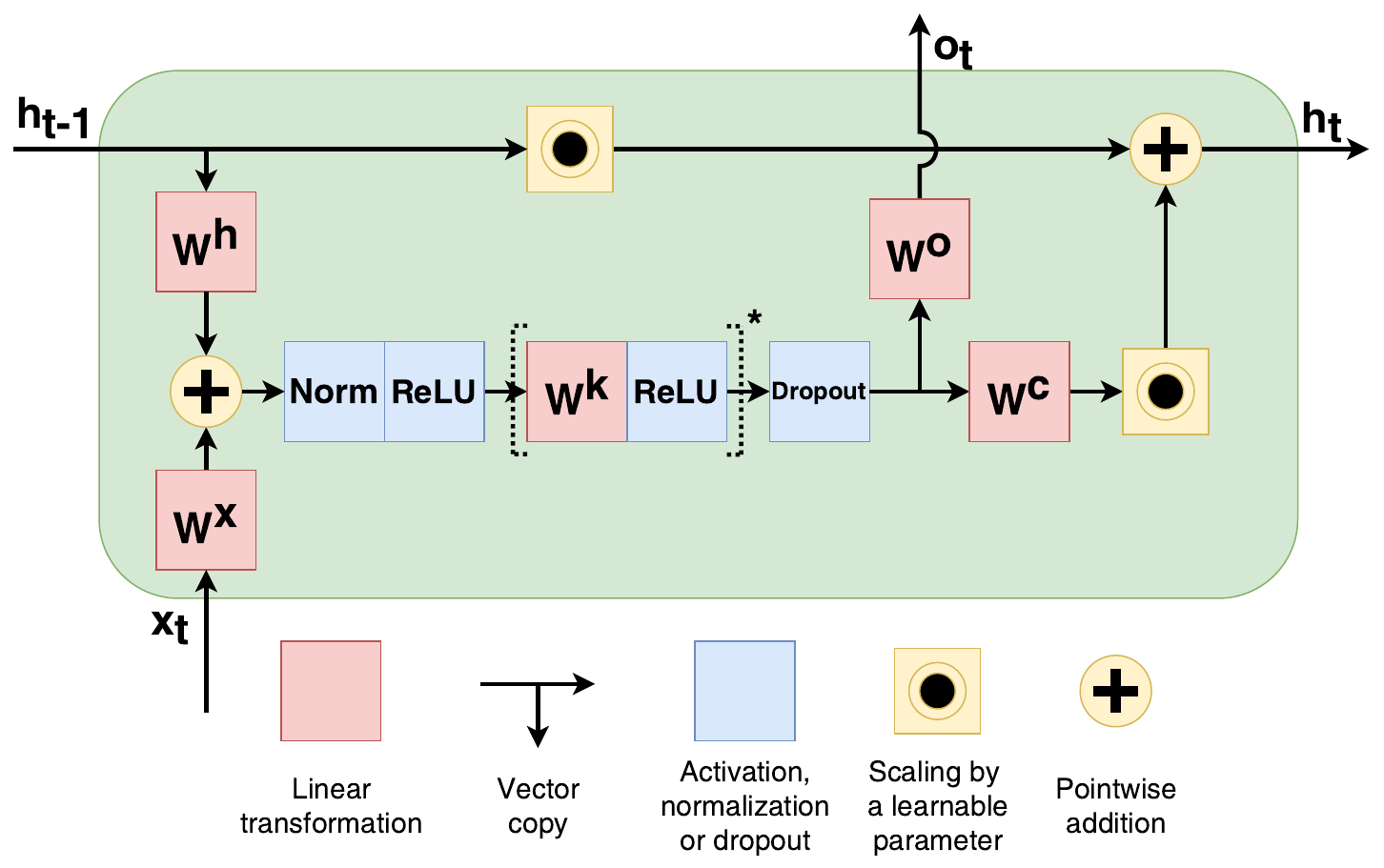}
\caption{The structure of the \textit{Residual Recurrent Unit} (RRU) with two state-input linear transformations. The dotted brackets cover the layer which can be used zero or more times, in this case – once.}
\label{fig:rru}
\end{center}
\end{figure}

In contrast, we show that gates are not essential at all to construct a well-performing recurrent unit. To this end, we develop a recurrent cell not containing a single gate (see the visualization in Figure~\ref{fig:rru}). The proposed cell surpasses not only GRU and LSTM but also the so-far best Mogrifier LSTM on many commonly used benchmark tasks. Our cell is based on a residual ReLU network employing normalization and ReZero \cite{bachlechner2020rezero} recurrent state update. If well-performing recurrent units without gates can be created, perhaps we shouldn't think of gates every time we think of recurrent networks and maybe this insight could help us create strong non-gated networks in the future.

\section{Relation to Prior Work}
\label{section:related}

A great number of RNN cells have been developed. They usually use gates to ensure stable training over many time steps. The most well-known is LSTM \cite{hochreiter1997long} which has an extra cell state that can pass information directly forward and three gates (forget, input, and output) that control the data flow. This specific structure has allowed it to become a state-of-the-art cell, still being used extensively to this day. Its success is considered to stem from the 3 gates at its core, and since then, it has been considered that RNN cells need gates for optimal performance.

Following the success of the LSTM, a new cell called GRU \cite{cho2014learning} was proposed, which is loosely based on LSTM and often referred to as the light version of the LSTM. This cell combines the forget and input gates of an LSTM, resulting in 2 gates total -- reset gate and update gate. As a lighter cell, it can be computed faster while the results are often similar.

Based on the marvelous results of LSTM and GRU, many extensions and modifications have been proposed. Phased LSTM \cite{neil2016phased} introduce a new time gate. Gating also in the depthwise direction is explored in \cite{yao2015depth,kalchbrenner2015grid}. The power of multiplicative interactions, which is the central operation in gates, is explored in \cite{krause2016multiplicative,wu2016multiplicative}. The number of gates is brought to the culmination in Mogrifier LSTM, which does input and hidden state modulation by multiple rounds of reset gates \cite{melis2019mogrifier}. Their results suggest that using about five rounds of reset gates gives the optimal performance and positions Mogrifier LSTM as the currently best RNN cell in WikiText-2, Penn Treebank word-level, and Penn Treebank character-level tasks.

Several works have questioned the necessity of the high number of gates in RNNs. \cite{greff2016lstm}, through numerous experiments, find that the forget gate and the output activation function are the most critical components of the LSTM block, and removing any of them impairs performance significantly. A similar conclusion was obtained in \cite{van2018unreasonable} and a new cell called JANET was proposed, which is based on the LSTM but uses just the forget gate. The minimalistic designs of recurrent cells with only one forget gate were proposed in \cite{zhou2016minimal,heck2017simplified}. The need for a reset gate in GRU was questioned in \cite{ravanelli2018light}, where they obtain improved accuracy in Automatic Speech Recognition with the unit having only an update gate, Batch Normalization, and ReLU activation.



There has been some tinkering done with using a residual shortcut connection in RNNs. Consequently, a Residual RNN \cite{yue2018residual} has been proposed. Res-RNN and gRes-RNN were proposed, where the former is a purely residual recurrent network and the latter combines residual shortcut connection with a gate. Their units show improved speed but are not able to unequivocally beat LSTM in terms of accuracy.  
\cite{kim2017residual} proposes a Residual LSTM, which is a mix between a regular LSTM and a residual shortcut connection to improve performance in case of many layers. Even with the addition of the residual shortcut connection, it still contains a high number of gates.

\section{Residual Recurrent Unit}
\label{section:rru}

We propose a new cell that does not contain a single gate but provides a competitive performance. We call this cell the \textit{Residual Recurrent Unit}, RRU for short. The cell consists of a residual shortcut connection and several linear transformations employing Normalization and ReZero \cite{bachlechner2020rezero} recurrent state update. ReZero is a simple zero-initialized parameter that controls how much of the residual branch contributes to the updated state. ReZero helps us make the unit structure gateless, which makes it more simple. As it is initialized as 0, only the previous state is taken into effect at the first few training steps, and the optimal weight for the residual branch emerges during training. ReZero ResNet has been shown to have a non-vanishing gradient yielding stable training for very deep convolutional networks and Transformers \cite{bachlechner2020rezero} and achieving great results. Here we adapt it to replace gates in recurrent networks to see what benefits it could give to recurrent units. 

The visualization of our cell can be seen in Figure~\ref{fig:rru}. RRU takes input $x_t$ at the current time-step $t$ which is a vector of dimension $m$ and the previous state $h_{t-1}$ of dimension $n$ and produces the updated state $h_t$ and the output $o_t$ of dimension $p$. The calculations done in the RRU, are described as:
\begin{gather}
j_t = \mathrm{ReLU}(\mathrm{Normalize}(W^{x}x_t+W^{h}h_{t-1}+b^j)) \label{eq:rru-eq1} \\
[j_t = \mathrm{ReLU}(W^{k}j_t+b^k)]^* \label{eq:rru-eq2} \\
d_t = \mathrm{Dropout}(j_t) \label{eq:rru-eq3} \\
c_t = W^{c}d_t+b^c \label{eq:rru-eq4} \\
h_t = \sigma(S) \odot h_{t-1} + Z\odot{c_t} \label{eq:rru-eq5} \\
o_t = W^{o}d_t+b^o \label{eq:rru-eq6}
\end{gather}

In the above equations, upper case letters depict the learnable parameters. Weight matrices with their respective dimensions are: $W^{x} \in \mathbb{R}^{m\times{g}}$; $W^{h} \in \mathbb{R}^{n\times{g}}$; $W^{k} \in \mathbb{R}^{{g}\times{g}}$; $W^{o} \in \mathbb{R}^{{g}\times{p}}$; $W^{c} \in \mathbb{R}^{{g}\times{n}}$, where $g$ is the hidden size which we choose to be equal to $q*(m + n)$, and $q$ is the middle layer size multiplier (we typically use values from 0.1 to 8.0). Bias vectors and their dimensions are: $b^{j} \in \mathbb{R}^g$; $b^{k} \in \mathbb{R}^g$, $b^{o} \in \mathbb{R}^p$; $b^{c} \in \mathbb{R}^n$. There are two learnable scaling factors $S$ and $Z$ of dimension $n$. The sigmoid function is denoted as $\sigma$, $[\cdot]^*$ denotes the use of the equation inside the brackets zero or more times, and $\odot$ denotes scalar multiplication.



The structure of RRU is similar to the ReZero residual network \cite{bachlechner2020rezero}. At first, the previous hidden state $h_{t-1}$ and the input of the current timestep $x_t$ are passed through a linear transformation followed by L2 normalization and ReLU. Then, zero or more linear transformations follow employing ReLU activation. To make the formulas clearer, Equation~\ref{eq:rru-eq2} has two vectors named $j_t$, in reality, each layer has its own $W^{k}$ and $B^{k}$. This value is a hyperparameter, for which each dataset has a specific value that works best, usually 1 or 2. Dropout  \cite{srivastava2014dropout} follows, and the result is linearly transformed to get two values – the output of the current timestep $o_t$ and the hidden state candidate $c_t$. The next hidden state $h_t$ is produced by a weighted sum of the previous state and the candidate. The candidate is scaled by a zero-initialized parameter Z according to the ReZero principle. There is a slight difference from the ReZero paper in that we use a separate scale for each feature map, but ReZero uses a common scaling factor for all maps, see Section~\ref{section:ablation} for evidence that our version works better. 

Another difference from the traditional ResNet architecture is that we have introduced a scaling factor $S$ for the residual connection (ResNet has a fixed $S=1$). Such change is motivated by the need for the recurrent network to forget some of the information from the previous time steps. The scale $S$ is limited to the range $0-1$ by the sigmoid function to eliminate unstable behavior in some cases, especially if $S$ has become negative. Since $S < 1$, the cell's memories gradually fade out. Note that is a learnable parameter (for each feature map separately) so the network can choose the rate of forgetting. Such fade-out is not so flexible as a forget gate employed in other cells but the network can compensate for it using the residual branch $c$. We have observed that the value of $S$ actually changes during training, mainly decreasing for most of the feature maps. This can be explained that the network initially uses the residual connection to provide a stable gradient for training but later learns to rely on the $c$ value more which provides greater control. The value of $S$ is initialized in the way that after the sigmoid, it is uniformly distributed in the range $0-1$. This idea is suggested by \cite{gu2020improving} and relieves us from the need for another hyperparameter that needs to be tuned. We also experimented with a constant initialization of $S$; that worked similarly but required an adequate constant, usually in the range $0.4-0.95$ depending on the dataset (see Section~\ref{section:ablation}). The ablation study also shows that $S$ is important in general; for Penn Treebank removing the scaling produces a significantly worse result.

The initial hidden state $h_0$ is prepared to have all zeros except the first feature map, which is set to $\frac{1}{4}\sqrt{n}$. Initializing with all zeroes causes a blowup due to the employed normalization in the case when zeros are given also as the input values $x_t$ in the first timesteps. Such inputs may occur if the input is padded by zeros from some shorter sequence.

Note that although Equation~\ref{eq:rru-eq5} resembles the update gate of LSTM, it is not because in the update gate, both of the multiplied values depend on the input or hidden state but in our case, $S$ and $Z$ are learnable parameters -- effectively constants after training. 

\section{Experiments}
\label{section:experiments}

To test our unit's performance against the chosen competitors (GRU, LSTM, Mogrifier LSTM), we run experiments on language modeling, music modeling, sentiment analysis, and MNIST image classification tasks. These tasks are commonly used for evaluating recurrent networks. All chosen units are equivalent in terms of trainable parameters in our experiments, to ensure a fair comparison.


For polyphonic music modeling, the task is to predict the next note when knowing the previous ones. The model's performance is measured by negative log-likelihood (NLL) loss which sums the negative logarithm of all the correct prediction probabilities. For this task, we chose four datasets that are usually evaluated together – JSB Chorales, Nottingham, MuseData, Piano-midi.de \cite{boulanger2012modeling}. They are already split into training, validation, and testing portions.
For word-level language modeling, we use the Penn Treebank dataset \cite{marcus1993building}. The aim of this dataset is to predict the next word by knowing the previous ones. The network's performance is measured in perplexity, which is the probability that a word will show up as the next word from the previous context.
For character-level language modeling datasets we chose enwik8 \cite{hutter2012human}, text8\footnote{\url{http://mattmahoney.net/dc/textdata}} and Penn Treebank \cite{marcus1993building}. The aim is to predict the next character from the previous ones. The performance is measured in BPC (bits per character), which is the number of bits used to represent a single character from the text.
Although word-level and character-level tasks may seem similar, they evaluate different aspects of the cell. The word-level task has a large vocabulary, so the emphasis is put on evaluating RRU's ability to deal with diverse inputs, but the character-level task has a large window size requiring the cell to remember and process long history.

The goal for the IMDB sentiment analysis dataset \cite{maas2011learning} is to predict the sentiment of the text as either positive or negative. This dataset is of interest because it is quite different from the previous ones. The performance of the network is measured as the accuracy of the predictions. The challenging part of the dataset is to learn the hidden state in a way that can understand even complicated structures, for example, double negatives, etc.

A frequently used dataset for evaluating recurrent networks is MNIST image classification. RNNs are not well suited for image classification; therefore, it puts every aspect of the network under stress. The aim here is to predict the number that each picture represents by processing a sequence that consists of all the pixels in the picture.

\begin{table*}[htbp]
\caption{Evaluation results on each task. The "+/-" notation means that we did a grid search through ten different dropout values to find the one which performs the best. Every metric over the horizontal line is better if it is lower and every metric under it is better if it is higher.}
\begin{center}
\begin{tabular}{ l  c  c  c  c  c }
\hline
\multicolumn{1}{c}{\multirow{2}{*}{Task}} & \multirow{2}{*}{Tuned} & \multirow{2}{*}{RRU} & \multirow{2}{*}{GRU} & \multirow{2}{*}{LSTM} & Mogrifier \\
 &  &  &  &  & LSTM \\
\hline
Music JSB Chorales (NLL) & + & \textbf{7.72} & 8.20 & 8.33 & 8.18 \\
Music Nottingham (NLL) & + & \textbf{2.92} & 3.22 & 3.25 & 3.28 \\
Music MuseData (NLL) & + & \textbf{7.03} & 7.31 & 7.24 & 7.22 \\
Music Piano-midi.de (NLL) & + & \textbf{7.38} & 7.58 & 7.54 & 7.52 \\
Word-level Penn Treebank (Perplexity) & + & \textbf{102.56} & 122.21 & 140.35 & 126.88 \\
Character-level Penn Treebank (BPC) & +/- & \textbf{1.27} & 1.28 & 1.34 & 1.29 \\
Character-level enwik8 (BPC) & - & 1.37 & 1.53 & 1.49 & \textbf{1.36} \\
Character-level text8 (BPC) & - & \textbf{1.35} & 1.55 & 1.44 & 1.37 \\
\hline
Sentiment analysis IMDB (Accuracy) & - & \textbf{87.20} & 87.04 & 85.89 & 86.23 \\
Sequential MNIST (Accuracy) & - & \textbf{98.74} & 98.36 & 92.88 & 98.14 \\
Permuted MNIST (Accuracy) & - & 97.67 & \textbf{98.68} & 97.39 & 97.81 \\
\hline
\end{tabular}
\end{center}
\label{tab:results}
\end{table*}


For music modeling and word-level language modeling, we run hyperparameter optimization on each cell, using Bayesian hyperparameter tuning from the HyperOpt library \cite{bergstra2013hyperopt}. The optimized hyperparameters include: the number of learnable parameters, dropout rate, learning rate, and some of the most influential hyperparameters specific to each cell; for more detailed configurations see Appendix~\ref{apx:configurations}. The RRU cell has a built-in dropout, so to get an equitable environment, we added recurrent dropout to GRU, LSTM, and Mogrifier LSTM as described in \cite{semeniuta2016recurrent}.
For character-level Penn Treebank, we also run optimization, but we only grid search through the different dropout rates from 0.0 to 0.9 -- and the best testing result from these runs is shown in the results table. This experiment is further described in Appendix~\ref{apx:dropout}.
For the rest of the datasets, hyperparameter tuning would require computational power exceeding our budget, so we only run a singular experiment on each cell on as fair a configuration as possible, that is, we try to use the same configuration but sometimes changes to the configuration have to be made, for more details see Appendix~\ref{apx:configurations}.

Each experiment was done on a single machine with 16 GB RAM and a single NVIDIA T4 GPU, time spent during a single, full training varies from averagely 4 minutes (JSB Chorales) to averagely 115 hours (enwik8). We use RAdam optimizer \cite{liu2019variance} with gradient clipping. We train each dataset until the validation loss stops decreasing for 3-11 epochs, depending on the dataset.

Results of all of these experiments can be seen in Table~\ref{tab:results}. RRU is a definite leader giving the best performance on almost all of the datasets. Notably, RRU scores on top for all fully hyperparameter-optimized cases. In the two cases where RRU is outperformed, it still comes close, and it is probable that RRU would achieve the best performance if full hyperparameter optimization were performed for these datasets. We suspect that the RRU could reach a state-of-the-art performance score on specific datasets, but we didn't have the time and resources to spend on tuning to achieve this. Convergence graphs of some of these experiments can be seen in Appendix~\ref{apx:convergence}, in which RRU usually converges first and with better results.

\section{Robustness against Hyperparameter Choice}
\label{section:robustness}

\begin{figure}[htbp]
\begin{center}
\includegraphics[width=0.8\linewidth]{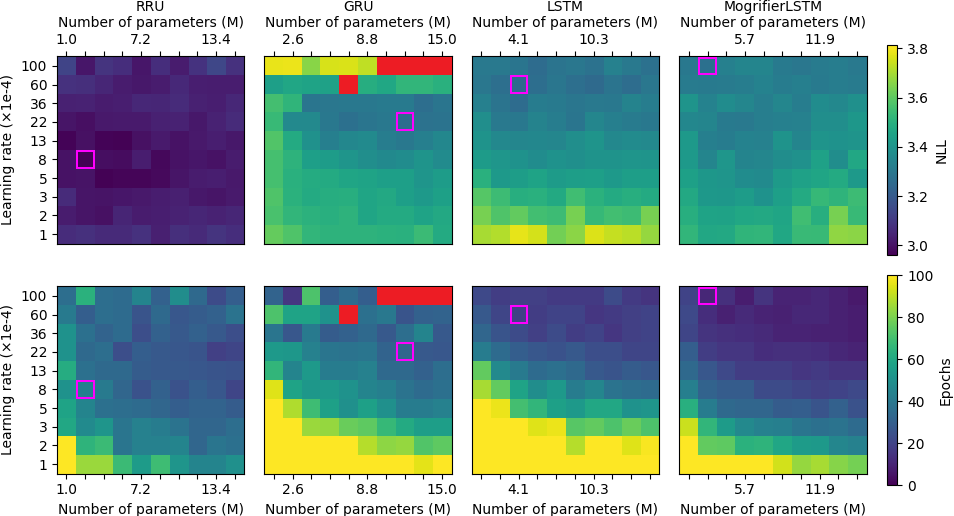}
\caption{Robustness against learning rate and parameter count on the Nottingham dataset. The top row gives the testing NLL, and the bottom row shows the number of epochs needed to reach the best NLL. Good results are depicted with blue color, and poor results -- with yellow. The red squares correspond to runs where the training failed, and the pink frames show the run with the best NLL. We observe that RRU consistently produces the best NLL across the entire range and achieves that within a small number of epochs.}
\label{fig:diversity-nottingham}
\end{center}
\end{figure}

One of the main reasons for LSTM's and GRU's wide usage is their robustness against hyperparameter choice; that is, usually, you can use them with some default parameters, and they will give decent results. Such property has been described in several works, one being \cite{talathi2015improving}.
We will test the robustness of each cell to the selection of learning rate and learnable parameter count -- the two most important hyperparameters. We run a grid search on the Nottingham music modeling dataset with a reasonably wide hyperparameter range for all the cells. We plot the validation NLL and the number of epochs needed to reach the best accuracy as heatmaps in Figure~\ref{fig:diversity-nottingham}. The figure shows that our cell has a much broader range of parameters in which it works well, which also means that our cell can be used successfully without much parameter tuning. Further robustness experiments can be seen in Appendix~\ref{apx:robustness}.

\section{Ablation Study}
\label{section:ablation}

In this section, we investigate the influence of the key elements in our cell by replacing them with simpler ones to determine whether our cell's structure is optimal. We compare the RRU performance against versions with:

\begin{itemize}
    \item Removed normalization.

    \item Using a single scalar multiplier $Z$ for all feature maps as proposed in the ReZero paper \cite{bachlechner2020rezero}. Our cell uses a different multiplier for each map.
    
    \item Added ReLU over Equations~\ref{eq:rru-eq4} and~\ref{eq:rru-eq6}. Such a setup is employed in some ResNet architectures \cite{he2016identity, he2016deep} and could have more expressive power.
    
    \item Residual weight $S$ set as a constant with a scalar value 1 i.e. using an unscaled residual shortcut connection, expressing Equation~\ref{eq:rru-eq5} as $h_t = h_{t-1} + Z\odot{c_t}$.
    
    \item Residual weight $S$ initialization with a scalar value 0.95 instead of a random one.
\end{itemize}

For these experiments, we have chosen character-level Penn Treebank, IMDB, Sequential MNIST datasets, on which we do a single run, and Nottingham, word-level Penn Treebank, on which we optimize the hyperparameters as described in Section~\ref{section:experiments}. For each dataset's configuration see the respective dataset's configuration for RRU in Section~\ref{section:experiments}, except for character-level Penn Treebank for which the configuration differs by the dropout rate which isn't tuned, but set as 0.3. The results can be seen in Table~\ref{tab:ablation}. We conclude that the current structure of our cell gives the best performance. Interestingly, the considered simplifications produce only slightly worse results, suggesting that the RRU's structure is robust to changes. Although great results can be achieved without normalization, it seems to give the RRU more stability during the training and contributes to robustness (see Appendix~\ref{section:robustness}). We notice that our approach of using a different ReZero multiplier for each hidden map gives better performance than using only the singular scalar ReZero parameter. Adding another ReLU over the vectors $c_t$ and $o_t$ does not help. Passing the full state without a scaling factor $S$ decreases performance. Using constant initialization seems to give a similar performance, but in these experiments, the random initialization gave us better results with the added bonus of fewer hyperparameters to tune.

\begin{table*}[htbp]
\caption{Ablation study results for all of the datasets. "Character-level Penn Treebank" is denoted as "Character PTB" and "Word-level Penn Treebank" is denoted as "Word PTB" to save space.}
\begin{center}
\begin{tabular}{ l  c  c  c c c}
\hline
& Character & \multirow{2}{*}{IMDB} & Sequential & \multirow{2}{*}{Nottingham} & Word \\
\multicolumn{1}{c}{Version} & PTB & \multirow{2}{*}{(Accuracy)} & MNIST & \multirow{2}{*}{(NLL)} & PTB \\
& (BPC) & & (Accuracy) & & (Perplexity) \\
\hline

\textbf{RRU} & \textbf{1.32} & \textbf{87.45} & \textbf{98.74} & \textbf{2.92} & \textbf{102.70} \\

no normalization & \textbf{1.32} & 87.21 & 98.03 & \textbf{2.92} & 104.60 \\

single scalar ReZero & \textbf{1.32} & 87.08 & 98.45 & 2.97 & 104.44 \\

ReLU over $c$ and $o_t$ & 1.34 & 87.17 & 86.38 & 3.07 & 120.76 \\

$S=1$ & 2.25 & 86.78 & 98.43 & 3.56 & 110.74 \\

$S$ initialized with $0.95$ & 1.33 & 86.71 & 98.60 & 2.96 & 103.09 \\

\hline
\end{tabular}
\end{center}
\label{tab:ablation}
\end{table*}

\section{Conclusion}
\label{section:conclusion}
By developing a new RNN cell without any gates and showing that it outperforms the gated cells on many tasks, we have demonstrated that gates are not necessary for RNNs. Our proposed RRU cell is robust to parameter selection and can be used in new tasks without much tuning. RRU has roughly the same speed as pure TensorFlow implementations of LSTM or GRU, and we look forward to low-level optimized CUDA RRU implementation that matches the speed of optimized LSTM and GRU implementations.
We expect that the insights gained in this work will contribute to further improvements in the designs of recurrent and residual networks.

\newpage
\appendix

\section{Appendix 1 – Dropout Analysis}
\label{apx:dropout}

\begin{figure}[htbp]
\begin{center}
\includegraphics[width=0.95\linewidth]{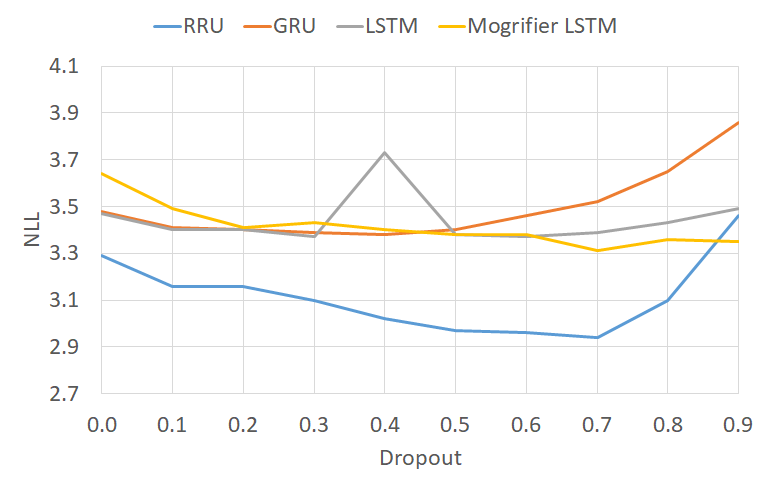}
\caption{NLL (lower is better) depending on the dropout rate for each cell on the Nottingham dataset.}
\label{fig:dropout-nottingham}
\end{center}
\end{figure}
\begin{figure}[htbp]
\begin{center}
\includegraphics[width=0.95\linewidth]{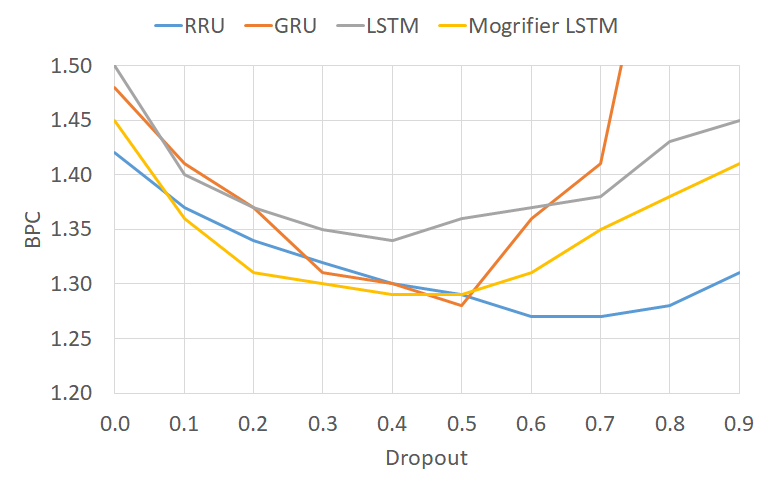}
\caption{BPC (lower is better) depending on the dropout rate for each cell on the character-level Penn Treebank dataset.}
\label{fig:dropout-ptbchar}
\end{center}
\end{figure}

While running experiments, we noticed that our cell seems to handle dropout better than the other cells. To test this observation, we ran a grid search on Nottingham and character-level Penn Treebank datasets for each cell through different dropout rates –- from 0.0 to 0.9, for more details of the configurations used see Appendix~\ref{apx:configurations}. The results from these experiments can be seen in Figures~\ref{fig:dropout-nottingham} and~\ref{fig:dropout-ptbchar}. From these results, we can see that all cells benefit from dropout, but for RRU, its impact is much more pronounced, and RRU ultimately reaches better final results than the other cells. We can also see that the RRU works best with a dropout rate of around 0.7. Possibly, dropout is better suited for ReLU networks, as is RRU, rather than for gated networks.

\clearpage

\section{Appendix 2 – Further Robustness Experiments}
\label{apx:robustness}

\begin{figure}[htbp]
\begin{center}
\includegraphics[width=0.95\linewidth]{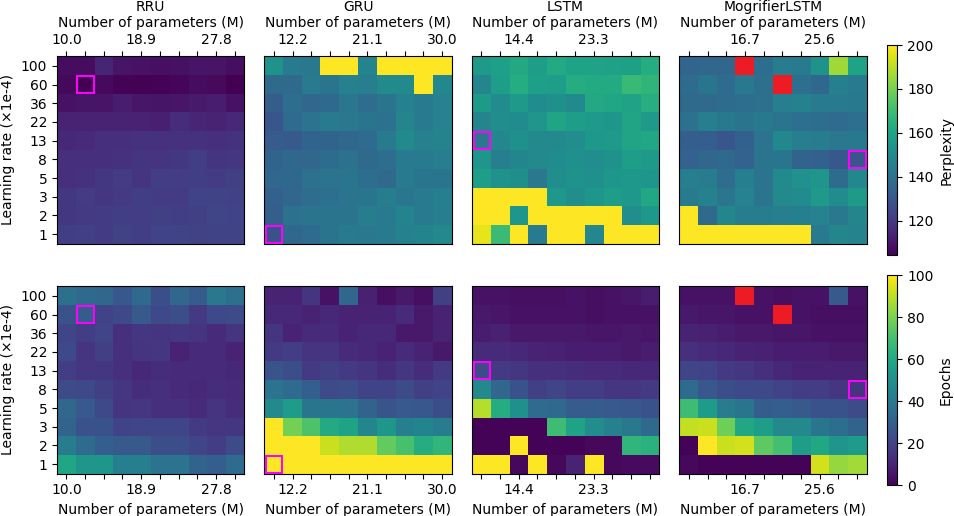}
\caption{Robustness against learning rate and parameter count for different cells on the word-level Penn Treebank dataset. The top row gives the testing perplexity, and the bottom row shows the number of epochs needed to reach the best perplexity. Good results are depicted with blue color, and poor results -- with yellow. The red squares correspond to runs where the training failed, and the pink frames show the run with the best perplexity. We observe that RRU consistently produces the best perplexity across the entire range and achieves that within a small number of epochs.}
\label{fig:diversity-ptbword}
\end{center}
\end{figure}

In Section~\ref{section:robustness} we tested the robustness of our cell. Here we will do the same experiment on a different dataset, the word-level Penn Treebank, to furthermore test the robustness. We plot the validation perplexity and the number of epochs needed to reach the best accuracy as heatmaps in Figure~\ref{fig:diversity-ptbword}. The figure shows that our cell has a much broader range of parameters in which it works well, which also means that our cell can be used successfully without much parameter tuning. We also notice that the LSTM and the Mogrifier LSTM had trouble learning anything at all on very small learning rates.

\begin{figure}[htbp]
\begin{center}
\includegraphics[width=0.95\linewidth]{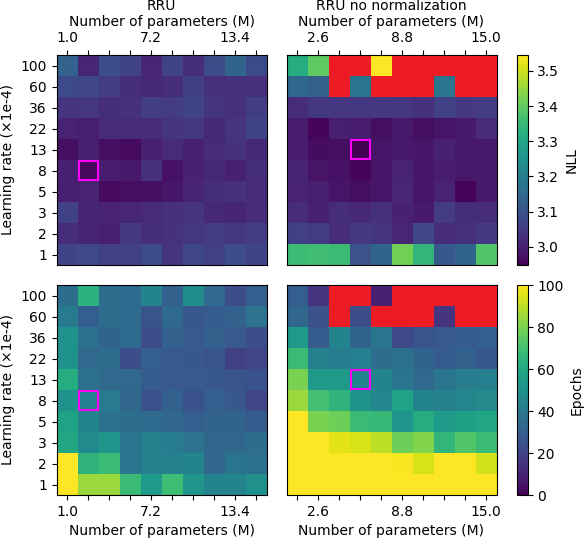}
\caption{Robustness against leaning rate and parameter count for different cells on the Nottingham dataset. The top row gives the testing NLL, and the bottom row shows the number of epochs needed to reach the best loss. Color depictions are the same as in Figure~\ref{fig:diversity-ptbword}.}
\label{fig:apx-c-diversity-rrubattle}
\end{center}
\end{figure}

\begin{figure}[htbp]
\begin{center}
\includegraphics[width=0.95\linewidth]{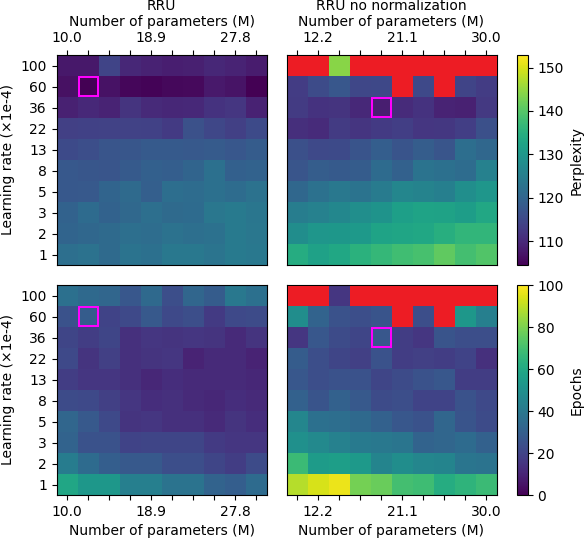}
\caption{Robustness against leaning rate and parameter count for different cells on the word-level Penn Treebank dataset. The top row gives the testing perplexity, and the bottom row shows the number of epochs needed to reach the best loss. Color depictions are the same as in Figure~\ref{fig:diversity-ptbword}.}
\label{fig:apx-c-diversity-rrubattle2}
\end{center}
\end{figure}

In Appendix~\ref{section:ablation} we looked at different RRU versions from which we concluded that our version tops the other versions, but we want to show in more detail why normalization is beneficial in the RRU. We will compare the RRU with and without normalization with the robustness tests as described in Section~\ref{section:robustness}. We plot the validation loss and the number of epochs needed to reach the best accuracy as heatmaps in Figure~\ref{fig:apx-c-diversity-rrubattle} and~\ref{fig:apx-c-diversity-rrubattle2}. We observe that RRU with normalization consistently produces the best loss across the entire range and achieves that within a small number of epochs which shows that normalization helps gain a much broader range of parameters in which it works well and helps avoid failed training sessions.

\clearpage

\section{Appendix 3 – Convergence Speed}
\label{apx:convergence}

Convergence speed is another important factor in neural networks, for that reason, we will display convergence graphs of three datasets in this section, one for each training mode – JSB Chorales (tuned), character-level Penn Treebank (half-tuned) and IMDB (not tuned). The graphs are taken from the experiments done in Section~\ref{section:experiments}.

\begin{figure}[htbp]
\begin{center}
\includegraphics[width=0.95\linewidth]{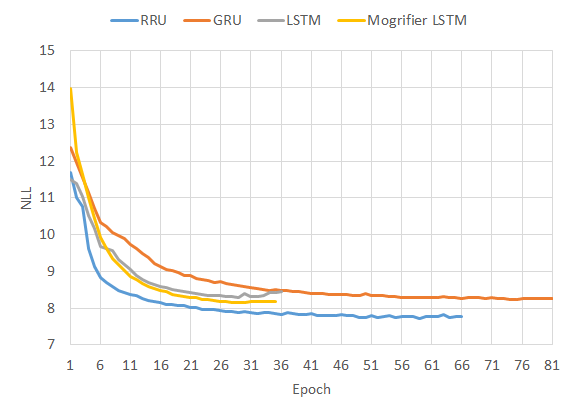}
\caption{Validation NLL on JSB Chorales dataset for each epoch and tuned cell.}
\label{fig:apx-b-jsb-chorales}
\end{center}
\end{figure}

The convergence graph of the JSB Chorales dataset with the best parameters found in tuning for each cell can be seen in Figure~\ref{fig:apx-b-jsb-chorales}. We can observe that RRU in this dataset trains much faster, beating every other cell's best NLL at the 17th epoch, which implies that early stopping could be done.

\begin{figure}[htbp]
\begin{center}
\includegraphics[width=0.95\linewidth]{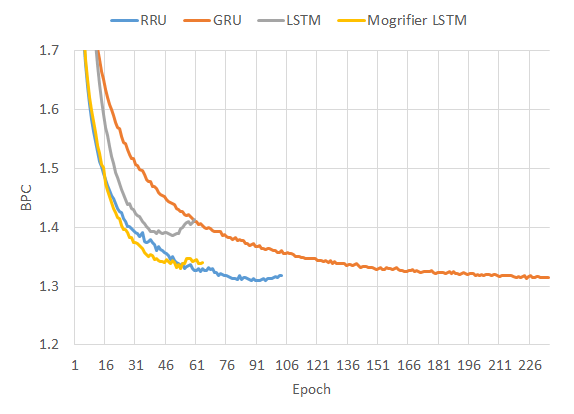}
\caption{Validation BPC on character-level Penn Treebank dataset for each epoch and half-tuned cell. In this Figure RRU has a 0.7 dropout, GRU has a 0.5 dropout, LSTM has a 0.4 dropout and Mogrifier LSTM has a 0.5 dropout.}
\label{fig:apx-b-ptb-char}
\end{center}
\end{figure}

The convergence graph of the character-level Penn Treebank dataset after half-tuning (going through 10 different dropout values and taking the best one) has been done can be seen in Figure~\ref{fig:apx-b-ptb-char}. We see that the RRU beats all the other cells here while doing it in a similar epoch count as others, even when it has a larger dropout rate, which usually increases epoch count tremendously.

\begin{figure}[htbp]
\begin{center}
\includegraphics[width=0.95\linewidth]{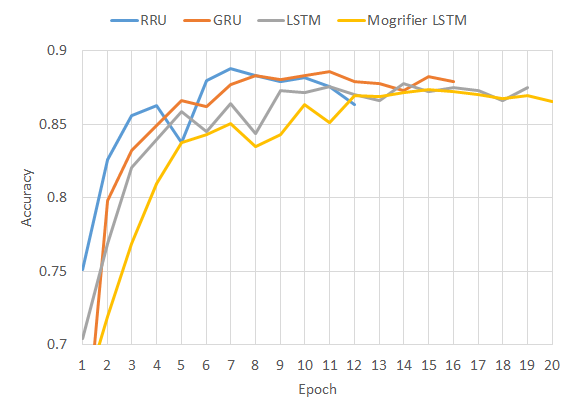}
\caption{Validation Accuracy on the IMDB dataset for each epoch and cell.}
\label{fig:apx-b-imdb}
\end{center}
\end{figure}

The convergence graph of the IMDB dataset can be seen in Figure~\ref{fig:apx-b-imdb}. In this graph, we once again see that the RRU is capable of reaching its best accuracy first and beating the other cells with it.

\clearpage

\section{Appendix 4 – Experiment Configurations}
\label{apx:configurations}

This appendix is intended for the elaboration on the configurations of the experiments that are written in Section~\ref{section:experiments}, but for specific configurations, we suggest contacting the author(s).

\subsection{Dataset Configurations}
\label{dataset-configuration}

Polyphonic music modeling datasets contain sequences of played piano-roll elements (MIDI note numbers between 21 and 108 inclusive), which we represent as a binary mask for each time step, where ones are placed in the positions corresponding to the MIDI notes played at that time step. We trim or pad the sequences to length 200, similarly to how it was done by \cite{subakan2017diagonal}.

Word-level Penn Treebank dataset is already split into training, validation and testing portions. We use a  context length of 64 and a vocabulary of 10 thousand, which includes all of the words in the dataset.

Character-level Penn Treebank also is already split into training, validation, and testing portions. However, enwik8 and text8 are not, so we use the standard split for each of them, which is 90\% training, 5\% validation, and 5\% testing. We use a window size of 512 for Penn Treebank and 256 for enwik8 and text8. The training is run in a stateful manner, meaning that we pass the old state forward to the next window (with a 10 \% chance of passing a state filled with zeros).

The IMDB dataset consists of 25 thousand training sequences and 25 thousand testing sequences. We take 5 thousand sequences from the end of the training set as a validation set. We trim the sequences to length 500 and use a vocabulary of size 10 thousand created from the most frequently occurring words (there are approximately 25 thousand different words in total) to be able to fit the sequences into our GPU memory.

We use the MNIST image classification dataset in a specific way: we use Sequential MNIST in which we take the image by pixel rows and get a single sequence and Permuted MNIST in which the pixels of each row are randomly permuted according to some fixed permutation, which makes the task harder. Each $28 \times 28$ image is transformed into a sequence of 784 elements. The dataset consists of 60 thousand training images and 10 thousand testing images. We reserve 10 thousand images from the training data for validation. 

\subsection{Main Experiments}
\label{main-configuration}

All four polyphonic music datasets use the same configuration. We set context size as 200 (as done in \cite{subakan2017diagonal}), batch size as 16, and the number of layers as 1, for RRU we use 1 ReLU layer and an output size of 64, all of these values we experimentally found to work well. The training stops when there hasn't been a lower NLL for 7 epochs.

For the word-level Penn treebank, we set context size as 64, batch size as 128, number of layers as 2, and embedding size of 64, for RRU we use 1 ReLU layer and output size of 128, all of these values we experimentally found to work well. The training stops when there hasn't been a lower perplexity for 7 epochs.

For polyphonic music modeling and word-level language modeling, we run hyperparameter optimization on each cell, where we tune the learning rate, the dropout rate, and the number of parameters in 100 runs, with ranges as described in Table~\ref{tab:apx-a-main-tuned}. We also tune the following parameters of the RNN cells: for RRU we tune the middle layer size multiplier $q$ in the range [0.1; 8.0]; for the LSTM we tune the forget bias in the range [-3.0; 3.0]; for the Mogrifier LSTM we tune the feature rounds in the range [5; 6] and the feature rank in the range [40; 90], as these were mentioned as the optimal ranges in the Mogrifier LSTM paper. Optimal values for the tuned parameters for each dataset and cell can be seen in Table~\ref{tab:apx-a-main-tuned-best}.

\begin{table*}[htbp]
\caption{Parameter ranges for main experiments that used tuning. $^u$ means the value is taken from the uniform scale and $^l$ means from the logarithmic scale. If '.' is present in the range, it means it includes float numbers, else the range is only from integers. 'M' means that the value is multiplied by $10^6$. These ranges were picked because they seemed sufficient to include all of the optimal values, which were approximately identified in the initial experiments.}
\begin{center}
\begin{tabular}{ c c c c }
\hline
Dataset & Learning rate & Dropout rate & Number of parameters \\
\hline
JSB Chorales & \multirow{5}{*}{$[1e-4; 1e-2]^l$} & \multirow{5}{*}{$[0.0; 0.8]^u$} & \multirow{4}{*}{$[1M - 15M]^u$} \\
Nottingham & & & \\
MuseData  & & & \\
Piano-midi.de & & & \\
\cline{4-4}
Word-level Penn Treebank & & & $[10M - 30M]^u$ \\
\hline
\end{tabular}
\end{center}
\label{tab:apx-a-main-tuned}
\end{table*}

\begin{table*}[htbp]
\caption{Optimal values for the main experiments that used tuning. 'M' means that the value is multiplied by $10^6$. "JSB Chorales" is denoted as "JSB", "MuseData" is denoted as "Muse", "Piano-midi.de" is denoted as "Piano" and "Word-level Penn treebank" is denoted as "Word PTB" to save space.}
\begin{center}
\begin{tabular}{ c c c c c c }
\hline
Parameter & JSB & Nottingham & Muse & Piano & Word PTB \\
\hline
\multicolumn{6}{c}{RRU} \\
\hline
Learning rate & 0.00495 & 0.00029 & 0.00030 & 0.00077 & 0.00683 \\
Dropout rate & 0.79362 & 0.77186 & 0.77518 & 0.71577 & 0.66035 \\
Number of parameters & 6.9M & 2.6M & 8.6M & 5.9M & 25.9M \\
Middle layer size multiplier $q$ & 1.76958 & 7.31994 & 6.85699 & 3.68371 & 6.86655 \\
\hline
\multicolumn{6}{c}{GRU} \\
\hline
Learning rate & 0.00460 & 0.00454 & 0.00260 & 0.00932 & 0.00012 \\
Dropout rate & 0.67605 & 0.33787 & 0.64839 & 0.75069 & 0.27062 \\
Number of parameters & 1.8M & 3.4M &  1.9M & 2.4M & 10.0M \\
\hline
\multicolumn{6}{c}{LSTM} \\
\hline
Learning rate & 0.01000 & 0.00882 & 0.00493 & 0.00727 & 0.00162 \\
Dropout rate & 0.30081 & 0.20310 & 0.35383 & 0.22951 & 0.10384 \\
Number of parameters & 1.3M & 1.3M & 1.8M & 11.9M & 14.2M \\
Forget bias & 0.76843 & -0.04104 & 1.85746 & 0.59993 & -2.33114 \\
\hline
\multicolumn{6}{c}{Mogrifier LSTM} \\
\hline
Learning rate & 0.00255 & 0.00169 & 0.00135 & 0.00012 & 0.00074 \\
Dropout rate & 0.47521 & 0.73808 & 0.55499 & 0.77291 & 0.46872 \\
Number of parameters & 1.3M & 6.0M & 10.3M & 2.9M & 24.2M \\
Feature mask rounds & 6 & 5 & 5 & 5 & 5 \\
Feature mask rank & 58 & 48 & 76 & 63 & 74 \\
\hline
\end{tabular}
\end{center}
\label{tab:apx-a-main-tuned-best}
\end{table*}

For character-level Penn Treebank we set the number of parameters as 24 million, learning rate as $1e-3$, we set context size as 512, batch size as 64, the number of layers as 2, and embedding size of 16, for RRU we use 2 ReLU layers, middle layer size multiplier $q$ as 4.0 and output size of 128, for LSTM we set forget bias as 1.0, for Mogrifier LSTM we use 4 feature mask rounds of rank 24, all of these values we experimentally found to work well. The training stops when there hasn't been a lower BPC for 11 epochs. We run this configuration through ten different dropout rates from 0.0 to 0.9 and report the one with the lowest BPC.

The remaining datasets – enwik8, text8, IMDB, Sequential MNIST, and P-MNIST – are run with no tuning. We tried to use the same configuration for each of the cells, but it wasn't always possible, because for some datasets some cells were unable to train with the common configuration. The configuration that was the same for each of the cells can be seen in Table~\ref{tab:apx-a-main-not-tuned-same} and the slight differences can be seen in Table~\ref{tab:apx-a-main-not-tuned-different}. All of these parameter values were experimentally found to work well. enwik8 number of parameters (48 million) were taken from the Mogrifier LSTM paper and Sequential MNIST, P-MNIST number of parameters (70 thousand) were taken from \cite{bai2018empirical}.

\begin{table*}[htbp]
\caption{Parameter values for parameters that are the same for each cell in non-tuned main experiments. "Sequential MNIST", and "P-MNIST" are denoted together as "MNISTs" and "text8", and "enwik8" are denoted together as "wiki8s". 'M' means that the value is multiplied by $10^6$. $[RRU]$ means the parameter is used in the RRU cell (the same denotation is used for the other cells).}
\begin{center}
\begin{tabular}{ c c c c c c }
\hline
Parameter & wiki8s & IMDB & MNISTs \\
\hline
Number of parameters & 48M & 20M & 70K  \\
Context size & 256 & 512 & 784 \\
Batch size & 64 & 64 & 64 \\
Number of layers & 2 & 2 & 2 \\
Embedding size & 32 & 64 & - \\
Breaks after ... epochs with no performance gain & 3 & 5 & 5 \\
$[RRU]$ ReLU layers & 2 & 1 & 1 \\
$[RRU]$ Middle layer size multiplier $q$ & 4.0 & 2.0 & 2.0 \\
$[RRU]$ Output size & 128 & 64 & 64 \\
$[LSTM]$ Forget bias & 1.0 & 1.0 & 1.0 \\
$[Mogrifier LSTM]$ Feature mask rounds & 6 & 5 & 6 \\
$[Mogrifier LSTM]$ Feature mask rank & 79 & 40 & 50 \\
\hline
\end{tabular}
\end{center}
\label{tab:apx-a-main-not-tuned-same}
\end{table*}

\begin{table*}[htbp]
\caption{Parameters that differ between cells on some datasets in non-tuned main experiments. '*' in the "Cell" field means that all cells have the same configuration.}
\begin{center}
\begin{tabular}{ c c c c c c }
\hline
Dataset & Cell & Learning rate & Dropout rate \\
\hline
enwik8 & * & 1e-3 & 0.7 \\
\hline
\multirow{4}{*}{text8} & RRU & 1e-3 & \multirow{4}{*}{0.7} \\
 & GRU & 1e-4 & \\
 & LSTM & 1e-3 & \\
 & Mogrifier LSTM & 1e-3 & \\
\hline
IMDB & * & 1e-3 & 0.5 \\
\hline
 & RRU & 1e-3 & 0.7 \\
Sequential MNIST & GRU & 1e-3 & 0.7 \\
\& P-MNIST & LSTM & 1e-4 & 0.7 \\
 & Mogrifier LSTM & 5e-5 & 0.5 \\
\hline
\end{tabular}
\end{center}
\label{tab:apx-a-main-not-tuned-different}
\end{table*}

\subsection{Robustness Experiments}

For Nottingham here we use the same configuration as the configuration for Nottingham in the main experiments (Section~\ref{main-configuration}), except no tuning is done, so we set the middle layer size multiplier $q$ as 2 and dropout rate as 0.5. There are 100 runs in total from a full grid search from learning rates $1e-4$, $2e-4$, $3e-4$, $5e-4$, $8e-4$, $13e-4$, $22e-4$, $36e-4$, $6e-3$, $1e-2$ (ten values from $1e-4$ to $1e-2$ in logarithmic scale) – and the number of parameters – 1.0M, 2.6M, 4.1M, 5.7M, 7.2M, 8.8M, 10.3M, 11.9M, 13.4M, 15.0M (ten values from 1.0M to 15.0M in uniform scale).

For word-level Penn Treebank here we use the same configuration as the configuration for word-level Penn Treebank in the main experiments (Section~\ref{main-configuration}), except no tuning is done, so we set middle layer size multiplier $q$ as 2 and dropout rate as 0.5. There are 100 runs in total from a full grid search from learning rates – $1e-4$, $2e-4$, $3e-4$, $5e-4$, $8e-4$, $13e-4$, $22e-4$, $36e-4$, $6e-3$, $1e-2$ (ten values from $1e-4$ to $1e-2$ in logarithmic scale) – and the number of parameters – 10.0M, 12.2M, 14.4M, 16.7M, 18.9M, 21.1M, 23.3M, 25.6M, 27.8M, 30.0M (ten values from 10.0M to 30.0M in uniform scale).

\subsection{Dropout Experiments}
\label{dropout-configuration}

For Nottingham here we use the same configuration as the configuration for Nottingham in the main experiments (Section~\ref{main-configuration}), except no tuning is done, so we set the number of parameters to 5 million, learning rate as $1e-3$, middle layer size multiplier $q$ as 2.0. We run this configuration through different dropout rates –- from 0.0 to 0.9.

For character-level Penn Treebank experiment configuration see character-level Penn Treebank experiment configuration in Section~\ref{main-configuration}.

\newpage

\bibliographystyle{splncs04}
\bibliography{refs}

\end{document}